\documentclass[11pt]{article}

\usepackage[final]{acl}

\usepackage{times}
\usepackage{latexsym}
\usepackage[T1]{fontenc}
\usepackage[utf8]{inputenc}
\usepackage{microtype}
\usepackage{inconsolata}
\usepackage{graphicx}
\usepackage{booktabs}
\usepackage{amsmath}
\usepackage{amssymb}
\usepackage{xcolor}

\title{Perturbation, Not Provenance: Why Activation Steering Does Not Make a Robust Text Watermark}

\author{
 \textbf{Sergey Pletenev\textsuperscript{1,2}}
\\
 \textsuperscript{1}AIRI,
 \textsuperscript{2}Skoltech
\\
\texttt{
\href{mailto:alex010rey@gmail.com}{alex010rey@gmail.com}
}}
\hypersetup{
  pdftitle={Perturbation, Not Provenance: Why Activation Steering Does Not Make a Robust Text Watermark},
  pdfauthor={Sergey Pletenev}
}

\begin{document}
\maketitle

\begin{abstract}
Linear directions in the activation space of large language models causally and universally influence model behavior. This suggests that they might serve as a basis for a robust text watermark. The idea is to steer the generation along a specific semantic direction and then recover the mark by probing the activations of the re-encoded text.
We test this hypothesis by conducting a series of experiments. We vary the strength of the steering and use four different directions (formality, sentiment, complexity, and a random control) across two different model families. We also test two paraphrasing attacks: a sentence-level one and a stronger paragraph-level one.
The results show that activation-steering watermarks are outperformed by a standard logit-based watermark (KGW) in terms of detection accuracy. At a matched level of fluency (within $1.15$ times the original text's perplexity), KGW achieved an AUC of $0.81$ against a sentence-level paraphraser, while the best steering direction only reached $0.65$. A stronger paragraph-level paraphraser lowered both ($0.65$ for KGW and $0.59$ for steering), narrowing the gap but not reversing it. On Qwen, it left the two indistinguishable but at near-chance levels for both. We further showed that the detectability provided by steering was not semantic: a random direction was just as detectable as a meaningful one. The signal was independent of which layer was probed and whether a surface feature control was used. We conclude that, in this case, semantic activation does not encode a localized, recoverable semantic mark, but rather a global, content-independent shift that can only be strongly detected when the text is degraded.
\end{abstract}

\section{Introduction}
\label{sec:intro}

Text watermarking is a process of embedding a detectable signal into machine-generated text, allowing a third party with a specific key to later verify the text's authenticity. A key aspect of watermarking is its resistance to paraphrasing, meaning that an adversary attempting to rewrite the text should not be able to easily remove the watermark. Watermarks based on logit functions, such as KGW \citep{kirchenbauer2023watermark}, bias token sampling towards a pseudo-random ``green list''. On the other hand, there are semantic watermarks, like SemStamp \citep{hou2024semstamp}, which target paraphrasing resistance by operating in the space of sentence embeddings.

A natural and previously unexplored alternative is to embed a watermark inside the forward pass of the model. Recent work has shown that linear directions in the residual stream can causally steer the behavior of the model and generalize across different contexts. \citep{turner2023activation,rimsky2024caa,zou2023representation,park2024linear}. If a semantic direction leaves a trace that survives paraphrasing because new sentences preserve meaning even when they rewrite tokens, then we could: (i) steer generation along a secret semantic direction and (ii) detect the watermark by re-encoding the candidate text through the base model and probing its activations. Such a watermark would be architecturally different from logit- and embedding-space methods and might inherit the robustness of the underlying semantic representation to paraphrasing.


We tested this hypothesis and reported a negative result. Our main contributions are:
\begin{enumerate}
\item We formalize and implement an activation-steering watermark and evaluate it against KGW on a detection-versus-fluency frontier, under both a sentence-level and a paragraph-level paraphrase attack, across steering strengths spanning fluent to degraded generation.
\item We show that the steering watermark is Pareto-dominated: at matched deployable fluency, it is never more detectable than KGW, and only reaches comparable detectability by inflating the perplexity by several-fold.
\item We demonstrate that its detectability is not semantic via three controls: a random steering direction matches meaningful ones, the signal is layer-invariant, and it cannot be explained by surface text statistics.
\end{enumerate}
We release data and code for further usage\footnote{\href{https://github.com/A1exRey/Steering-Text-Watermarking}{https://github.com/A1exRey/Steering-Text-Watermarking}}.

\section{Related Work}
\label{sec:related}
 
\paragraph{Logit-based watermarks.}
KGW \citep{kirchenbauer2023watermark} and successors \citep{zhao2024provable} partition the vocabulary and
bias sampling; detection is a statistical test over token counts. These methods
are known to degrade under paraphrase, motivating semantic approaches.
 
\paragraph{Semantic / paraphrase-robust watermarks.}
SemStamp \citep{hou2024semstamp} and its variants \citep{hou2024ksemstamp} operate in sentence-embedding
space via rejection sampling to survive paraphrase. Our method differs by
intervening in the residual stream; we find this confers no robustness advantage
over the logit baseline at matched fluency.
 
\paragraph{Activation steering.}
Linear directions in activation space causally and generalizably affect model
behavior \citep{turner2023activation,rimsky2024caa,zou2023representation,park2024linear}, which motivated the watermarking hypothesis we test. Our negative result delimits that promise: such directions do not yield a localized, recoverable semantic channel that survives paraphrase.

Concurrent work, SLAM \citep{harelcanada2026slamstructurallinguisticactivation}, instantiates an activation-injection watermark using sparse-autoencoder structural directions (voice, tense, clause order) on Gemma-2. Our results delimit this specific case -- the directions that transmit are behavioral and therefore collide with natural text. Steered structure that is not lexically anchored is what a paraphrasing tool removes. We test a sentence-level paraphrasing tool only, so we do not claim to overcome structural marks in SLAM; rather, we describe the secrecy-robustness tension that any such system must navigate.

\section{Method}
\label{sec:method}
 
\paragraph{Watermark by activation steering.}
Let $h^{(\ell)}_t \in \mathbb{R}^d$ be the residual-stream activation at layer
$\ell$ and token position $t$. Given a unit-norm steering direction
$v \in \mathbb{R}^d$, we add, at every position during generation,
\begin{equation}
    \tilde{h}^{(\ell)}_t = h^{(\ell)}_t + \alpha \, \lVert h^{(\ell)}_t \rVert \, v,
    \label{eq:steer}
\end{equation}
where $\alpha$ is a \emph{norm-relative} steering fraction. Scaling by
$\lVert h^{(\ell)}_t \rVert$ makes $\alpha$ comparable across models and layers
whose residual norms differ by orders of magnitude (e.g.\ $\approx\!10^2$ for
Llama-3.2-1B at $\ell{=}8$ versus $\approx\!2.2\!\times\!10^3$ for Qwen3-1.7B at
$\ell{=}14$); an absolute offset that steers one model is negligible or
destructive for another.
 
\paragraph{Detection.}
The verifier re-encodes candidate text through the unmodified base model,
mean-pools the layer-$\ell$ activations over non-padding positions, and trains a
regularized logistic probe to separate steered from unsteered text. We report
detection as the cross-validated ROC-AUC of this probe (pooled across generation
seeds).
 
\paragraph{Directions.}
We extract steering directions by mean-difference of pooled activations over
contrastive text pairs \citep{rimsky2024caa} generated by a larger model of the same family: formality
(formal vs.\ informal), sentiment (positive vs.\ negative), and complexity
(complex vs.\ simple). As a control we use a \emph{random} unit direction
orthogonalized against formality. If steering encodes a semantic mark, the
meaningful directions should be detectable and the random one should not.
 
\paragraph{Baselines.}
 
We compare against KGW \citep{kirchenbauer2023watermark} as implemented in, sweeping its logit bias $\delta$. KGW is detected with with its native $z$-score test, not the activation probe.
For paraphrasing we use two attackers. The first rewrites every generation sentence-by-sentence
with a Pegasus \citep{zhang2020pegasus}. Because a sentence-level paraphraser is not able to reorder or merge material across sentence boundaries, we also use DIPPER \citep{krishna2023paraphrasing}, a $11B$ paragraph-level paraphraser that rewrites in few sentence windows.
 
\section{Experimental Setup}
\label{sec:setup}
 
\paragraph{Models.}
We use Llama-3.2-1B-Instruct \citep{dubey2024llama3} and Qwen3-1.7B \citep{qwen2025qwen3} as generators, with the 8B model of each family used only to synthesize contrastive pairs.
 
\paragraph{Prompts and generation.}
We sample $200$ Alpaca instruction prompts and generate with temperature $1.0$, top-$p$ $0.9$ and $150$ max new tokens. Every operating point is generated under $3$ seeds ($600$ generations per class).
 
\paragraph{Steering grid.}
The norm-relative fraction $\alpha$ is swept per model over a grid chosen by calibration to span fluent-but-changed to degraded generation ($\alpha \in \{0.1, 0.2, 0.4, 0.6, 0.8\}$). Because all figures plot detection against \emph{measured} perplexity rather than against $\alpha$, per-model grids remain directly comparable in fluency space.
 
\paragraph{Metrics.}
In both frontier figures (Figures~\ref{fig:pareto-llama} and~\ref{fig:pareto-qwen}) and in Table~\ref{tab:frontier}, the $y$-axis is post-paraphrase detection AUC; we report symmetric AUC ($\max(\text{AUC}, 1{-}\text{AUC})$) so a key-aware verifier's separability is never under-counted. For the $x$-axis we use a \emph{two-sided distortion} 
$d = \lvert \log(\text{ppl}_{\text{steered}}/\text{ppl}_{\text{unsteered}}) \rvert$, 
the absolute log of the perplexity ratio under the base model. A one-sided "fluency cost" (perplexity ratio) is not adequate here because, on Qwen, several steering directions lower perplexity relative to the unsteered baseline, we analyze this in Results section.
 
\section{Results}
\label{sec:frontier}
 
\begin{table}[t]
\centering
\small
\setlength{\tabcolsep}{4pt}
\begin{tabular}{lcccc}
\toprule
Methods & Ppl. & $d$ & AUC$_{\text{Peg}}$ & AUC$_{\text{DIP}}$ \\
\midrule
\multicolumn{5}{l}{\emph{Deployable regime} ($\leq 1.15\times$ ppl)} \\
KGW ($\delta{=}3$)         & $1.06$ & $0.06$ & $\mathbf{0.81}$ & $\mathbf{0.65}$ \\
sentiment ($\alpha{=}0.1$) & $1.03$ & $0.03$ & $0.65$ & $0.59$ \\
random ($\alpha{=}0.2$)    & $1.13$ & $0.12$ & $0.61$ & $0.57$ \\
complexity ($\alpha{=}0.4$)& $1.03$ & $0.03$ & $0.59$ & $0.55$ \\
formality ($\alpha{=}0.2$) & $1.05$ & $0.05$ & $0.57$ & $0.52$ \\
\midrule
\multicolumn{5}{l}{\emph{Steering's high-detectability points}} \\
formality ($\alpha{=}0.6$) & $1.73$ & $0.55$ & $0.91$ & $0.83$ \\
sentiment ($\alpha{=}0.2$) & $1.32$ & $0.28$ & $0.89$ & $0.83$ \\
random ($\alpha{=}0.4$)    & $1.98$ & $0.68$ & $0.86$ & $0.78$ \\
\bottomrule
\end{tabular}
\caption{Llama-3.2-1B, best point per method within each case, under the
Pegasus (Peg) and DIPPER (DIP) attacks. Ppl.\ is the ratio to unsteered and
$d=\lvert\log(\text{ppl ratio})\rvert$}
\label{tab:frontier}
\end{table}
 
\begin{figure}[t]
\centering
\includegraphics[width=\columnwidth]{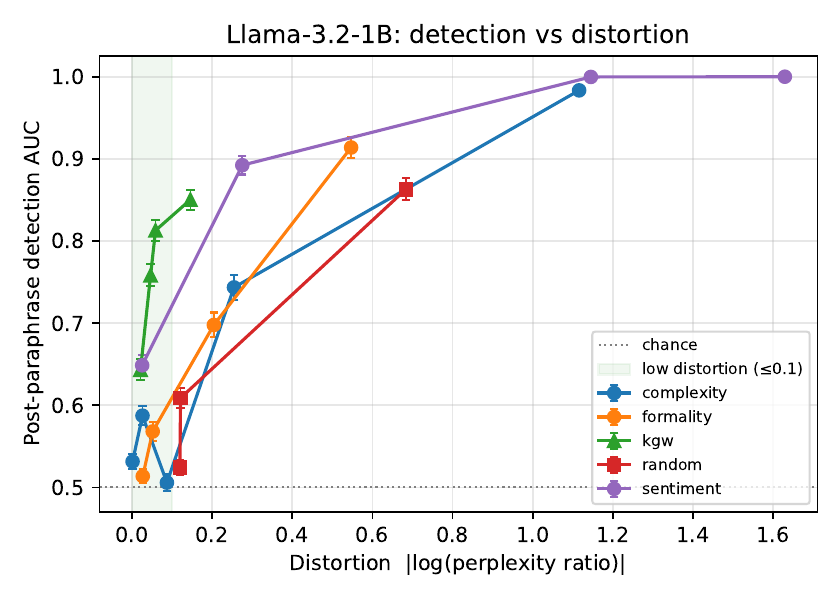}
\caption{Detection-versus-distortion frontier (Llama-3.2-1B)}
\label{fig:pareto-llama}
\end{figure}

\begin{figure}[t]
\centering
\includegraphics[width=\columnwidth]{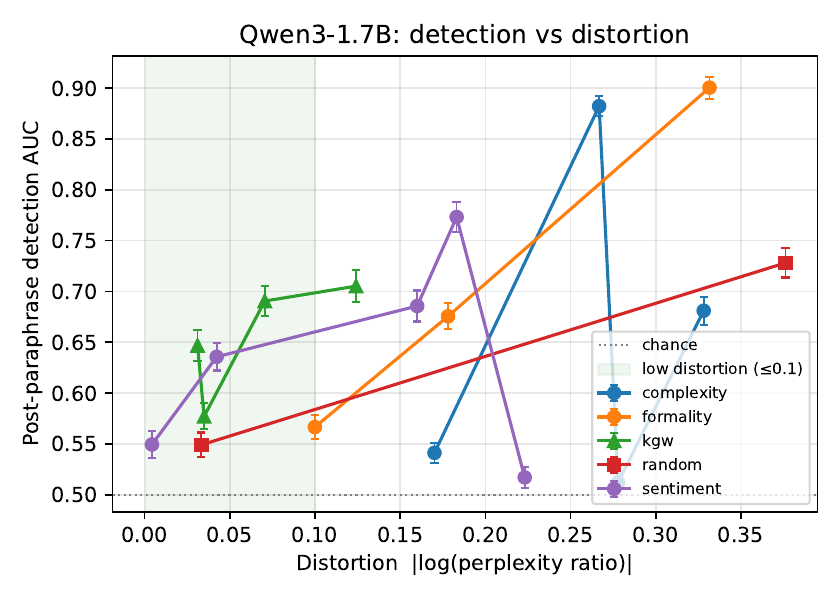}
\caption{Detection-versus-distortion frontier (Qwen3-1.7B)}
\label{fig:pareto-qwen}
\end{figure}

Table~\ref{tab:frontier} and Figures~\ref{fig:pareto-llama} and~\ref{fig:pareto-qwen} show the main results. In the deployable case, KGW achieves $0.81$ post-paraphrase AUC with a perplexity of $1.06$. No other approach comes close to this level of performance at comparable fluency. Sentiment achieves the best result with $0.65$ AUC, and the remaining three approaches fall between $0.57$ and $0.61$. Steering matches KGW's best operating point ($0.85$) only from a perplexity of $1.16$ upward. Non-sentiment approaches only perform well at $1.7$-$2\times$, which corresponds to a noticeable degradation in text quality. The logit watermark dominates the activation-steering watermark across the frontier, except for one case: on Qwen, sentiment with $\alpha=0.8$ reaches $0.77$, which is higher than KGW's best ($0.71$), but at a greater distortion ($d=0.18$ compared to $0.12$). 
 

Because both detectors score the same unsteered negative set, their AUCs are correlated, so we bootstrap the \emph{difference} directly, resampling whole prompts (each prompt contributes one generation per seed to each class) and recomputing both AUCs on every replicate. In the deployable band KGW exceeds the best steering direction by $0.16$ AUC under Pegasus, and no replicate put the difference at or below zero. The curves do not overlap within their error bars anywhere in the low-distortion regime (Figure~\ref{fig:pareto-llama}).

\paragraph{A stronger attacker narrows the gap.}
DIPPER is much more damaging to KGW than Pegasus. When the attack is turned on, KGW loses an average of $0.15$ AUC on Llama and $0.11$ AUC on Qwen over its four different bias settings, while the four $\alpha=0.1$ points lose only $0.01$ AUC, an order of magnitude less on both. This is expected, as KGW is more susceptible to the attack than the other models.

A paragraph-level rewriter disrupts the token-level green list, whereas a probe on mean-pooled activations reads a global perturbation that partly survives rewording. On Llama, the gap therefore falls from $0.16$ AUC under Pegasus to $0.06$ under DIPPER, so Pegasus alone is an artifact of a weak attacker.
On Qwen, the ordering does not survive: the best in-band steering point (sentiment, $\alpha{=}0.4$)) reaches $0.587$ against KGW's $0.557$, a difference of $-0.03$ with interval $[-0.07, +0.01]$. Under the stronger attack, the two are statistically indistinguishable on Qwen, and we do not claim KGW wins there. However, Qwen is the model on which KGW performs the worst, so the convergence reflects a decreasing baseline as well as an increasing steering curve. $0.587$ is close enough to the chance level to be interpreted as neither method being effective here, rather than the steering curve succeeding.

At the top of the steering grid, generation degenerates into empty or repetitive text whose perplexity ratio exceeds $10\times$ the unsteered baseline. We exclude these points from all figures and tables: they are not deployable watermarks but destroyed text.


\paragraph{Steering can reduce perplexity}

On Qwen3-1.7B several steering directions produce a perplexity ratio below $1$: steered generations are \emph{more} probable under the base model than unsteered ones (formality and complexity at $\alpha \le 0.4$ give $0.72$--$0.90$). Steering towards "formal" or "complex" registers did not degrade fluency so much as homogenized output towards higher-probability, more generic phrasing.

This gives us a methodological implication: detectability rises even as perplexity falls, so it cannot be a function of "fluency cost" alone. What the detector tracked was displacement from the base distribution in either direction, which was why our figures used the two-sided axis $d$. Under it, the apparent "free detection" of these points resolved -- they sat at moderate distortion ($d = 0.1$--$0.33$, not at zero cost.

A second implication was that perplexity alone could not decide which points were deployable. Strong steering failed predominantly by \emph{repetition}, and repeated tokens are highly predictable, so a collapsed completion has a \emph{low} perplexity. On Qwen, this puts formality $\alpha{=}0.6$ at a $1.15\times$ ratio with $0.98$ AUC -- apparently a deployable watermark beating KGW -- when one completion in ten has actually collapsed into a repetition loop. We therefore exclude operating points where more than 5\% of completions are empty or have fewer than half their word 4-grams unique. On Qwen, this removes formality $\alpha{=}0.6$ and complexity $\alpha{=}0.8$; on Llama, it removes nothing that the perplexity rule hadn't already removed, so Table~\ref{tab:frontier}. Appendix~\ref{app:degeneracy} gives the criterion, an independent cross-model check confirming the effect is not an artifact of self-scoring, and per-point collapse rates. Detectability is achieved through degeneration, rather than a semantic marker.

\section{Ablation study}
\label{sec:mechanism}
 
If the steering encoded a semantic watermark, three predictions would hold: a meaningful direction would be more detectable than a random one, the signal would be strongest at the steered layer, and detection would not reduce to surface text statistics. However, the first two predictions fail outright, and the third is not refuted, but it is ineffective in the context that is relevant. At deployable fluency, there is no signal to explain surface statistics, because there is barely any signal at all. Where detection appears, it appears together with distortion, and it is available to a random direction as readily as it is to a meaningful one.

\paragraph{Random matches semantic.}
At matched steering strength the random control is among the most detectable directions, not the least. At $\alpha{=}0.4$ it reaches $0.86$ AUC on Llama and $0.97$ on Qwen, above formality ($0.70$ and $0.90$), above complexity ($0.59$ and $0.68$), and on Qwen above sentiment ($0.64$) as well; only Llama sentiment exceeds it. A direction carrying no semantic content is therefore no harder to recover than one carrying a concept.

Direction identity does not even transfer between models. Sentiment is the most detectable direction on Llama at every steering strength and the \emph{least} detectable on Qwen, though both were extracted by the same mean-difference procedure from the same contrastive template. 

\begin{figure}[t]
\centering
\includegraphics[width=\columnwidth]{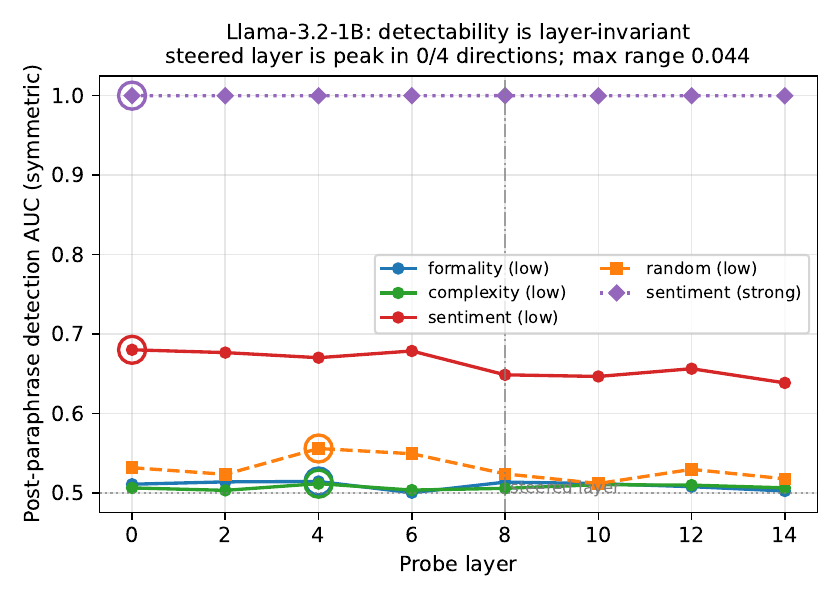}
\caption{Post-paraphrase detectability against probe layer on Llama-3.2-1B. Rings
mark each direction's peak layer; the dash-dotted line is the layer that was
steered. "low" is $\alpha{=}0.1$ and "strong" is $\alpha{=}0.4$.}
\label{fig:layer}
\end{figure}

\begin{figure}[t]
\centering
\includegraphics[width=\columnwidth]{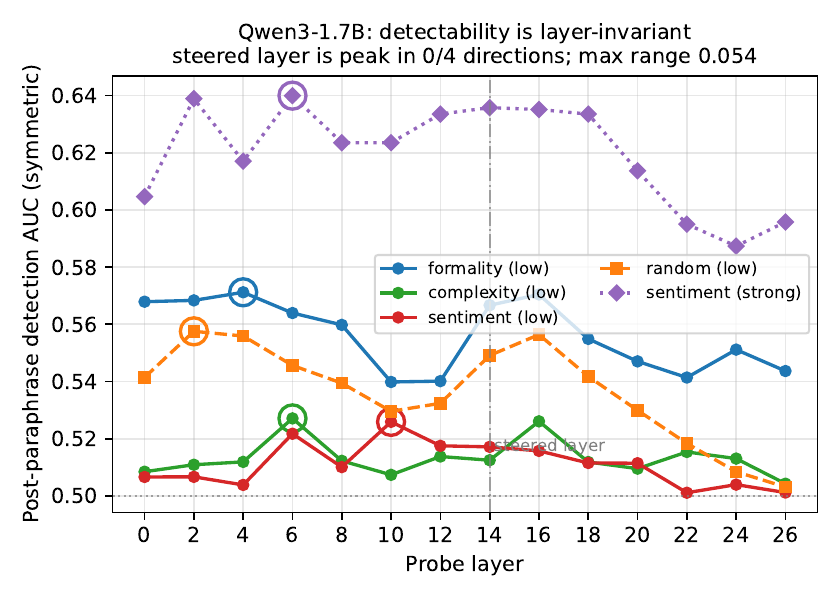}
\caption{Post-paraphrase detectability against probe layer on Qwen3-1.7B}
\label{fig:layer-qwen}
\end{figure}

\paragraph{Layer invariance.}
The signal is flat across layers. We probe every other layer across the full depth of each model and compare detectability. Post-paraphrase AUC varies by at most $0.044$ (Llama, $8$ layers) and $0.054$ (Qwen, $14$ layers) across all probed layers, and the layer we actually steered is the probe-AUC maximum in none of the four directions on either model ($0/4$ each); the peaks land at scattered, unrelated layers. A watermark localized to the steered representation would peak there. Instead the signal is a global property of the generated text, detectable at any depth (Figures~\ref{fig:layer} and~\ref{fig:layer-qwen}). Because we cover the whole stack rather than a few layers, this is not an undersampling artifact, and because the pattern is the same in both model families it is not a configuration or anomaly of one architecture.

\paragraph{Surface statistics.}

At low, fluent steering ($\alpha{=}0.1$) neither probe separates the two sets. A probe trained only on shallow features (length, type--token ratio, punctuation rates) reaches $0.51$--$0.56$ AUC, and the activation probe reaches $0.51$--$0.65$; the two agree to within $0.02$ for every direction on both models except sentiment on Llama, the single point where steering is detectable at deployable fluency. Matching the two sets on completion length leaves the activation AUC essentially unchanged, so length is not the confound either. The correct reading is not that a deep signal survives a surface control, but that at fluent steering there is almost no signal to begin with.

At strong steering ($\alpha{=}0.4$) both probes rise together, and the surface probe becomes accurate exactly where the text is most distorted ($0.90$ for the random direction on Qwen, $0.81$ for sentiment on Llama). Surface detectability therefore rises in lockstep with distortion for the random control just as for the meaningful directions, which is the same conclusion the other two controls reach: detection tracks displacement from the base distribution, not content.

 
\paragraph{The sentiment exception.}
Sentiment is the one direction that becomes strongly detectable at modest perplexity on Llama (AUC $0.89$ at $1.32\times$). It is, however, also layer-invariant: heavy sentiment steering makes the entire text conspicuously more positive, a global shift detectable at any layer rather than a localized mark. Nor does it replicate---on Qwen the same direction is the weakest of the four at every steering strength, reaching only $0.77$ even at $\alpha{=}0.8$. The exception is a property of one direction in one model, and it illustrates the mechanism rather than contradicting it.
 
\section{Conclusion}
\label{sec:conclusion}

Semantic activation-steering watermarks are an interesting idea, but they do not work as expected. In the testing regime we used, which included two model families with approximately 1-2 billion parameters, a single steered layer at mid-depth, and both sentence-level and paragraph-level paraphrasing attacks, a logit watermark always dominated the model, regardless of the fluency cost. What we found was that what steering provided was non-semantic: a random direction matched meaningful ones, the signal was layer-invariant, and what survived a surface-statistics control was content-independent.

Steering encoded a global perturbation that was only strongly detectable when it degraded text. We don't claim that this rules out all activation-space watermarks -- structural directions or larger models may behave differently -- but it limits the semantic version of the idea, and we hope this saves others from taking a detour.
 
 
\section*{Limitations}
We compare against a logit-based watermark (KGW) but not against a sentence-embedding watermark such as SemStamp \citep{hou2024semstamp}. That is arguably the more natural comparison: like ours, and unlike KGW, it is designed for paraphrase robustness by operating on meaning rather than on tokens. 

We studied models with up to $\approx$2B parameters and steered a single mid-depth layer in each model. Larger models, where steering directions may be cleaner and more linear, could potentially change the picture. However, this is beyond the scope of this study, and it is possible that our setup tests a less favorable case for the steering hypothesis than a large-scale study would, so the negative result is unlikely to change with scale.

The concern about steering a single layer is also addressed by our design: the detector probes every layer and the signal is global rather than limited to the steered layer. Therefore, the claim does not depend on the specific layer being steered. Our detector is a deliberately strong key-aware probe; a more realistic key-blind detector would only lower steering detectability, reinforcing the negative result.

\bibliography{custom}

\appendix

\section{Appendix}
\label{app:main}

\subsection{Full results}
\label{app:points}

Tables~\ref{tab:all-llama} and~\ref{tab:all-qwen} list every cases behind Figures~\ref{fig:pareto-llama} and~\ref{fig:pareto-qwen}, so the figures can be read numerically. They also make explicit something the figures cannot show: on Qwen the perplexity ratio is \emph{not} monotone in the steering fraction, so the horizontal position of a point does not follow $\alpha$. For complexity, $\alpha{=}0.1$ sits at $d{=}0.28$ while $\alpha{=}0.6$ sits at $d{=}0.27$; a line joining points in distortion order therefore visits the steering grid out of sequence. On Llama the ordering is preserved.

\begin{table}[h]
\centering\small
\setlength{\tabcolsep}{4pt}
\begin{tabular}{llcccc}
\toprule
Method & $\alpha/\delta$ & Ppl. & $d$ & AUC$_{\text{Peg}}$ & AUC$_{\text{DIP}}$ \\
\midrule
unsteered   & --  & $1.00$  & --      & $0.50$ & $0.50$ \\
\midrule
formality   & $0.1$ & $1.03$ & $0.03$ & $0.514$ & $0.536$ \\
            & $0.2$ & $1.05$ & $0.05$ & $0.568$ & $0.519$ \\
            & $0.4$ & $1.23$ & $0.20$ & $0.698$ & $0.589$ \\
            & $0.6$ & $1.73$ & $0.55$ & $0.914$ & $0.831$ \\
            & $0.8$ & $11.9$ & $2.47$ & $0.999^{\dagger}$ & $0.999^{\dagger}$ \\
\midrule
complexity  & $0.1$ & $1.09$ & $0.09$ & $0.506$ & $0.505$ \\
            & $0.2$ & $1.00$ & $0.00$ & $0.532$ & $0.513$ \\
            & $0.4$ & $1.03$ & $0.03$ & $0.587$ & $0.550$ \\
            & $0.6$ & $1.29$ & $0.25$ & $0.744$ & $0.644$ \\
            & $0.8$ & $3.05$ & $1.12$ & $0.983$ & $0.969$ \\
\midrule
sentiment   & $0.1$ & $1.03$ & $0.03$ & $0.649$ & $0.593$ \\
            & $0.2$ & $1.32$ & $0.28$ & $0.892$ & $0.825$ \\
            & $0.4$ & $3.14$ & $1.15$ & $1.000$ & $0.998$ \\
            & $0.6$ & $5.10$ & $1.63$ & $1.000$ & $1.000$ \\
            & $0.8$ & $12.0$ & $2.48$ & $1.000^{\dagger}$ & $1.000^{\dagger}$ \\
\midrule
random      & $0.1$ & $1.13$ & $0.12$ & $0.524$ & $0.508$ \\
            & $0.2$ & $1.13$ & $0.12$ & $0.609$ & $0.568$ \\
            & $0.4$ & $1.98$ & $0.68$ & $0.863$ & $0.780$ \\
            & $0.6$ & $13.8$ & $2.63$ & $0.979^{\dagger}$ & $0.964^{\dagger}$ \\
            & $0.8$ & $86.9$ & $4.46$ & $0.996^{\dagger}$ & $0.990^{\dagger}$ \\
\midrule
KGW         & $1$ & $1.02$ & $0.02$ & $0.643$ & $0.558$ \\
            & $2$ & $1.05$ & $0.05$ & $0.758$ & $0.592$ \\
            & $3$ & $1.06$ & $0.06$ & $\mathbf{0.813}$ & $\mathbf{0.652}$ \\
            & $4$ & $1.16$ & $0.15$ & $0.850$ & $0.659$ \\
\bottomrule
\end{tabular}
\caption{Llama-3.2-1B, all operating points, under the Pegasus (Peg) and DIPPER
(DIP) attacks. Bold marks the best KGW point inside the deployable band. $\dagger$ excluded from the frontier by the perplexity rule.}
\label{tab:all-llama}
\end{table}

\begin{table}[h]
\centering\small
\setlength{\tabcolsep}{4pt}
\begin{tabular}{llcccc}
\toprule
Method & $\alpha/\delta$ & Ppl. & $d$ & AUC$_{\text{Peg}}$ & AUC$_{\text{DIP}}$ \\
\midrule
unsteered   & --  & $1.00$  & --      & $0.50$ & $0.50$ \\
\midrule
formality   & $0.1$ & $0.90$ & $0.10$ & $0.567$ & $0.534$ \\
            & $0.2$ & $0.84$ & $0.18$ & $0.676$ & $0.592$ \\
            & $0.4$ & $0.72$ & $0.33$ & $0.900$ & $0.772$ \\
            & $0.6$ & $1.15$ & $0.14$ & $0.985^{\dagger}$ & $0.956^{\dagger}$ \\
            & $0.8$ & --     & --     & $0.987^{\dagger}$ & $0.977^{\dagger}$ \\
\midrule
complexity  & $0.1$ & $0.76$ & $0.28$ & $0.512$ & $0.521$ \\
            & $0.2$ & $0.84$ & $0.17$ & $0.541$ & $0.545$ \\
            & $0.4$ & $0.72$ & $0.33$ & $0.681$ & $0.588$ \\
            & $0.6$ & $0.77$ & $0.27$ & $0.882$ & $0.768$ \\
            & $0.8$ & $2.37$ & $0.86$ & $0.980^{\dagger}$ & $0.956^{\dagger}$ \\
\midrule
sentiment   & $0.1$ & $1.25$ & $0.22$ & $0.517$ & $0.536$ \\
            & $0.2$ & $1.00$ & $0.00$ & $0.550$ & $0.550$ \\
            & $0.4$ & $1.04$ & $0.04$ & $0.636$ & $0.587$ \\
            & $0.6$ & $1.17$ & $0.16$ & $0.686$ & $0.677$ \\
            & $0.8$ & $1.20$ & $0.18$ & $0.773$ & $0.732$ \\
\midrule
random      & $0.1$ & $0.97$ & $0.03$ & $0.549$ & $0.507$ \\
            & $0.2$ & $1.46$ & $0.38$ & $0.728$ & $0.670$ \\
            & $0.4$ & $5834$ & $8.67$ & $0.967^{\dagger}$ & $0.930^{\dagger}$ \\
            & $0.6$ & --     & --     & $0.995^{\dagger}$ & $0.982^{\dagger}$ \\
            & $0.8$ & $3549$ & $8.17$ & $0.999^{\dagger}$ & $0.990^{\dagger}$ \\
\midrule
KGW         & $1$ & $1.04$ & $0.03$ & $0.577$ & $0.518$ \\
            & $2$ & $1.03$ & $0.03$ & $0.647$ & $0.532$ \\
            & $3$ & $1.07$ & $0.07$ & $0.691$ & $0.554$ \\
            & $4$ & $1.13$ & $0.12$ & $\mathbf{0.705}$ & $\mathbf{0.557}$ \\
\bottomrule
\end{tabular}
\caption{Qwen3-1.7B, all operating points, under the Pegasus (Peg) and DIPPER
(DIP) attacks. Ratios below $1$ are homogenization, not error. ``--'' marks
generations too degenerate to score. Bold marks the best KGW point inside the deployable band. $\dagger$ excluded from the frontier by the perplexity rule. }
\label{tab:all-qwen}
\end{table}

\subsection{The degeneracy criterion}
\label{app:degeneracy}

\paragraph{Why perplexity is insufficient.}
The frontier already discards points whose perplexity ratio exceeds $10\times$ as destroyed text. That rule cannot catch repetition, because a looping completion is highly predictable and therefore scores \emph{low} perplexity. A point can be visibly destroyed and still sit in the deployable band. Qwen formality $\alpha{=}0.6$ is such a point: $1.15\times$ perplexity, $0.98$ AUC, and completions such as "\emph{The singular and singularly singular singular singular and singularly singular and singularly singular and\ldots}"

\paragraph{This is not an artifact of self-scoring.}
Steering toward "formal" homogenizes text toward the steered model's own high-probability phrasing, so one might suspect the ratio is depressed because the scorer is the model that was steered. It is not. Re-scoring every operating point under an independent Llama-3.1-8B judge places the same point at $1.13\times$, against $1.15\times$ self-scored. Both scorers agree, because the effect is a property of the text: no perplexity based axis can rank output whose dominant failure mode is one that perplexity rewards.


\end{document}